\documentclass[letterpaper, 10 pt, conference]{ieeeconf}  
\IEEEoverridecommandlockouts                              
\usepackage{graphics} 
\usepackage{epsfig} 
\usepackage{mathptmx} 
\usepackage{times} 
\usepackage{amssymb}  
\usepackage[acronym,nomain,nogroupskip,style=superborder,nonumberlist,shortcuts]{glossaries}
\usepackage{algorithm}
\usepackage{algorithmic}
\usepackage{amsmath}
\usepackage{capt-of}   
\usepackage{lipsum}    
\usepackage{cite}
\usepackage{afterpage}
\usepackage{multirow}
\usepackage{adjustbox}
\usepackage{placeins}
\usepackage{color}
\usepackage{graphicx}
\usepackage{amsmath,amsfonts,amssymb}
\usepackage{cite}
\usepackage{array}
\usepackage{soul}
\usepackage{color}
\usepackage{graphics} 
\usepackage{epsfig} 
\usepackage{mathptmx} 
\usepackage{times} 
\usepackage{amsmath} 
\usepackage{amssymb}  
\usepackage{hyperref}
\usepackage{booktabs}
\usepackage{adjustbox}
\usepackage{float}
\usepackage[T1]{fontenc}
\usepackage{multirow}
\usepackage{paralist, tabularx}
\usepackage[dvipsnames]{xcolor}
\usepackage{soul}
\usepackage[normalem]{ulem}

\newacronym{rl}{RL}{Reinforcement Learning}
\newacronym{ssl}{SSL}{Self-Supervised Learning}
\newacronym{ours}{TAR}{Teacher-Aligned Representations via Contrastive Learning}
\newacronym{pomdp}{POMDP}{Partially Observable Markov Decision Process}
\newacronym{id}{ID}{In-Distribution}
\newacronym{ood}{OOD}{Out-of-Distribution}

\title{\LARGE \bf
\acs{ours}: \acrlong{ours} for Quadrupedal Locomotion
}

\author{Amr Mousa$^{*,1}$, Neil Karavis$^{2}$, Michele Caprio$^{1}$, Wei Pan$^{1}$ and Richard Allmendinger$^{1}$ 
\thanks{$^{1}$Amr Mousa, Richard Allmendinger, Wei Pan, and Michele Caprio are with the University of Manchester, United Kingdom.}%
\thanks{$^{2}$Neil Karavis is with BAE Systems, United Kingdom.}%
\thanks{* Project website: https://amrmousa.com/TARLoco/}
}

\makeatletter
\newcommand{\titlefigure}{%
  \begin{center}
    \refstepcounter{figure}
    \addtocounter{figure}{-1}%
    \includegraphics[width=\linewidth,height=80pt]{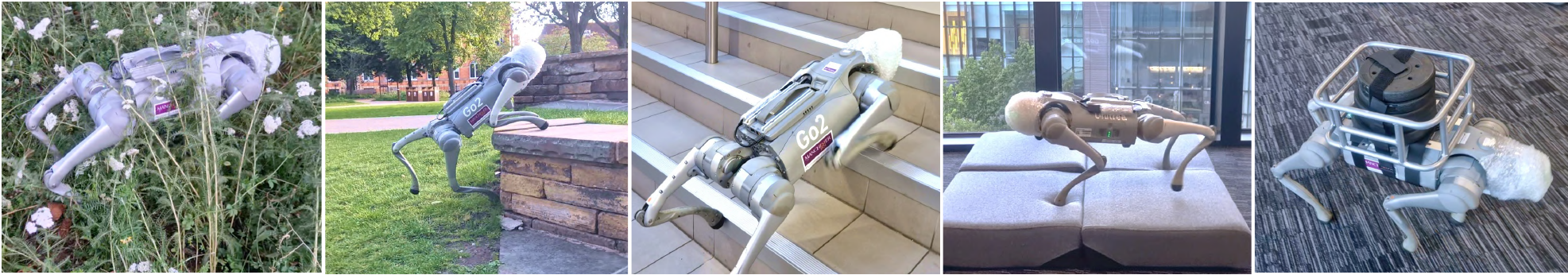}\\[-6pt]
    {\footnotesize                    
     \setlength{\baselineskip}{8.5pt}
     \captionof{figure}{%
       Real-world deployment of our locomotion policy on the Unitree Go2 robot across diverse scenarios: dense vegetation, high-step descent, stair climbing, soft foam traversal, and 10kg payload transport. Supplementary videos are available \href{https://amrmousa.com/TARLoco/}{online}\textsuperscript{*}.%
     }\label{fig:deployment}}%
  \end{center}\vspace{-1.25em}
}
\apptocmd{\@maketitle}{\titlefigure}{}{}
\makeatother

\begin{document}
\maketitle
\thispagestyle{empty}
\pagestyle{empty}
\addtocounter{figure}{-1}%

\begin{abstract}
Quadrupedal locomotion via \gls{rl} is commonly addressed using the teacher-student paradigm, where a privileged teacher guides a proprioceptive student policy. However, key challenges such as representation misalignment between privileged teacher and proprioceptive-only student, covariate shift due to behavioral cloning, and lack of deployable adaptation; lead to poor generalization in real-world scenarios. 
We propose \gls{ours}, a framework that leverages privileged information with self-supervised contrastive learning to bridge this gap. 
By aligning representations to a privileged teacher in simulation via contrastive objectives, our student policy learns structured latent spaces and exhibits robust generalization to \gls{ood} scenarios, surpassing the fully privileged “Teacher”. 
Results showed accelerated training by 2× compared to state-of-the-art baselines to achieve peak performance. \gls{ood} scenarios showed better generalization by 40\% on average compared to existing methods. 
Moreover, \gls{ours} transitions seamlessly into learning during deployment without requiring privileged states, setting a new benchmark in sample-efficient, adaptive locomotion and enabling continual fine-tuning in real-world scenarios. Open-source code and videos are available at \url{https://amrmousa.com/TARLoco/}.
\end{abstract}

\section{Introduction}

Recent advances in \acrfull{rl} have revealed a new era of data-driven quadrupedal locomotion, enabling robots to autonomously learn agile and adaptive motion skills \cite{lee2020overchallenge,kumar2021rma, miki2022wild, kim2023not, xiao2024paloco, chen2024slr,  margolis2022rapidlocomotion,long2024him}. \gls{rl}-based policies exhibit emergent behaviors that generalize beyond the training distribution, demonstrating adaptability in non-stationary settings.

The quadrupedal locomotion problem is naturally formulated as a \gls{pomdp}, where the robot makes decisions under incomplete state information. Due to limited sensor feedback, the robot cannot perceive critical environmental parameters such as ground friction, restitution, and external perturbations. Similarly, intrinsic robot state parameters, including payload shifts, mass, and linear velocity, remain uncertain, making locomotion control particularly challenging \cite{lee2020overchallenge, kumar2021rma, chen2024slr}. Effectively handling this partial observability requires a structured learning approach that enables the policy to infer unobserved state variables from available sensory inputs.

To address this challenge, teacher-student frameworks have been widely adopted to leverage privileged information during training while ensuring real-world deployability.  In this paradigm, a teacher policy is first trained with privileged information, leveraging exteroceptive sensing and latent state variables to construct a feature space that encodes environment and dynamics patterns. A student policy, trained concurrently \cite{wang2024cts} or in a secondary stage \cite{lee2020overchallenge, kumar2021rma, miki2022wild, kim2023not, xiao2024paloco}, learns to mimic the teacher’s behavior using only proprioceptive sensing through supervised learning.

However, this approach poses fundamental challenges. Without exteroceptive inputs (e.g., vision or depth sensing), the student must rely on proprioceptive feedback to infer missing information, a significant limitation in tasks requiring long-horizon context (e.g., stepping over obstacles or traversing uneven terrain). To improve temporal modeling, prior works have introduced MLPs with $N$-step observation history \cite{chen2024slr,long2024him,wang2024cts,peng2024bipedawalk,kumar2021rma,margolis2023walk},
along with memory-augmented architectures such as GRUs \cite{kim2024dyansti}, LSTMs \cite{siekmann2020bipedallstm,miki2022wild,wu2023amp}, and TCNs \cite{li2024bipedalversatile,lee2020overchallenge}. Although these methods enhance state retention, they remain constrained by the absence of exteroceptive and often fail alone to reconstruct the necessary latent representations for unseen scenarios beyond training.

Another challenge is \emph{representation misalignment} between teacher and student policies. The teacher’s latent space leverages
privileged observations, while the student—trained via behavioral cloning that regresses to a similar feature space using fewer modalities.
This mismatch impairs real-world generalization. Moreover, behavioral cloning aggravates covariate shift:
small discrepancies in the teacher’s trajectory accumulate during deployment, degrading performance \cite{Tennenholtz2021CovariateShift}.

Real-world deployment then suffers from the \emph{Sim2Real gap} \cite{ajanid2023domrand,vuong2019pickdomainrandomizationparameters, chen2024slr,lee2020overchallenge,long2024him}, as simulated dynamics can deviate from actual physics. Domain randomization helps by perturbing simulation parameters, but remains within limited ranges due to practical considerations \cite{ajanid2023domrand,dulacarnold2019realrl}. Lastly, fine-tuning on physical platforms is hindered by the lack of privileged information, restricting both \emph{long-term autonomy} and policy adaptation. 


These limitations motivate our work, which utilizes representation learning to bridge the gap between privileged and proprioceptive-only policies. Instead of naïve feature regression, we introduce a structured teacher-aligned contrastive learning method that enables the student policy to construct robust task-relevant latent spaces, enhancing generalization.

\vspace{3mm}
Our key contributions include:
 
\begin{itemize}
    \item Efficient Representation Alignment: We propose a contrastive teacher-aligned method that leverages privileged information to guide self-supervised representation learning. By aligning proprioceptive latent spaces without direct teacher-student regression, our method mitigates representation misalignment. It improves sample efficiency, achieving its peak performance in 50\% less training time compared to state-of-the-art baselines while attaining a higher return. This structured learning approach enhances generalization, reducing \acrfull{ood} evaluation error by 42.2\% compared to existing methods.

    \item Robust Adaptation and Negative Sampling: Our model incorporates a task-informed negative sampling strategy that improves representation learning, contributing a 8\% boost in evaluation metrics. Additionally, privileged information enhances performance by 28.2\%, reinforcing its role in learning robust policies.

    \item Deployable and Off-Policy Compatible Learning: Our approach eliminates reliance on privileged observations post-simulation by aligning student representations through contrastive learning. The teacher-aligned latent spaces integrate seamlessly with off-policy \gls{rl}, enabling real-world fine-tuning and continual adaptation.
\end{itemize}

In this paper, section~\ref{sec:related_work} reviews \gls{ssl} techniques in \gls{rl}, focusing on contrastive learning and representation alignment. It also draws connections between our approach and existing methods. 
Section~\ref{sec:methodology} introduces our framework, while section~\ref{sec:results} analyzes the performance by benchmarking against state-of-the-art methods and conducting ablation studies. 
Finally, Section~\ref{sec:conclusion} highlights key findings and outlines directions for future research.
\section{Related Work}
\label{sec:related_work}

\begin{figure}[t]
  \centering
  \fboxsep=3pt 
  \fboxrule=0pt 
  \fbox{
    \begin{minipage}{0.47\textwidth} 
      \centering
      \includegraphics[width=\textwidth]{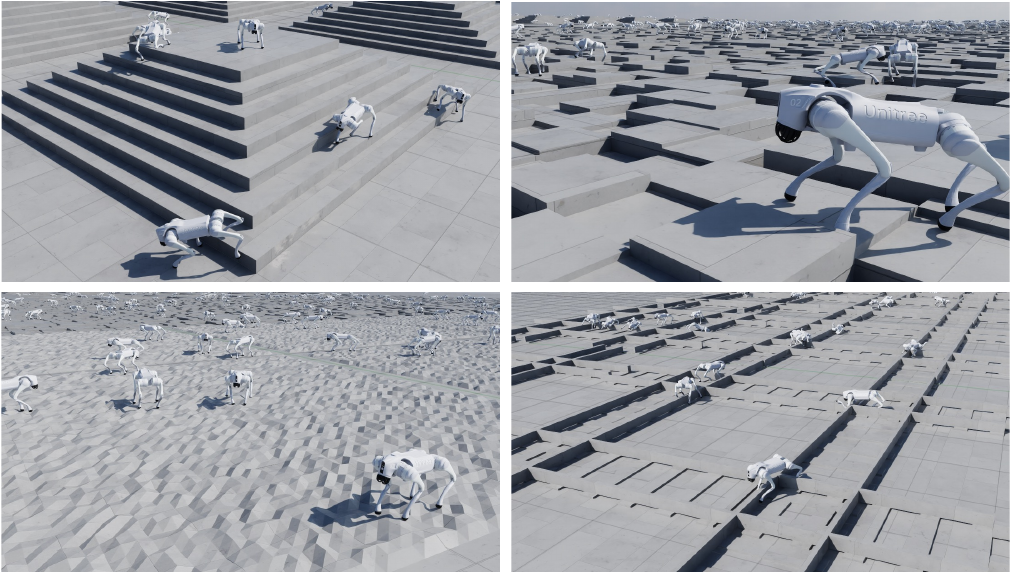} 
      \label{fig:terrain_in_sim}
    \end{minipage}
  }
  \caption{Generated terrains for training and testing, adapted from \cite{lee2020overchallenge}. We extend this setup by introducing challenging rail crossings with steep 25 cm steps, encouraging the robot to develop more robust locomotion strategies.}  
\label{terrain_figure}
  \label{terrain_figure}
\end{figure}

In this section, we review two key areas related to our work: \gls{ssl} and metric learning. Both play a significant role in shaping robust representations for decision-making in \gls{rl}.

\subsection{\acrlong{ssl}}

\gls{ssl} aims to derive meaningful latent representations from unlabeled data by leveraging the inherent structure within inputs. A widespread \gls{ssl} strategy in \gls{rl} is to reconstruct raw observations \cite{nahrendra2023dreamwaq,miki2022wild, yang2019dataefficientreinforcementlearning} to better model the underlying task dynamics. However, reconstructing high-dimensional states can lead models to capture irrelevant noise and redundant details \cite{deng2021dreamerpro}, reducing their focus on the most decision-critical features. Moreover, regression-based objectives in such setups can overfit to simulator-specific artifacts \cite{tiboni2023dropo} and cause latent space collapse—challenges that become more severe when domain randomization is used to handle real-world uncertainties.

\begin{figure*}[t]
  \centering
  \fboxsep=4pt 
  \fboxrule=0pt 
  \fbox{
    \begin{minipage}{0.9\textwidth} 
      \centering
      \includegraphics[width=\textwidth]{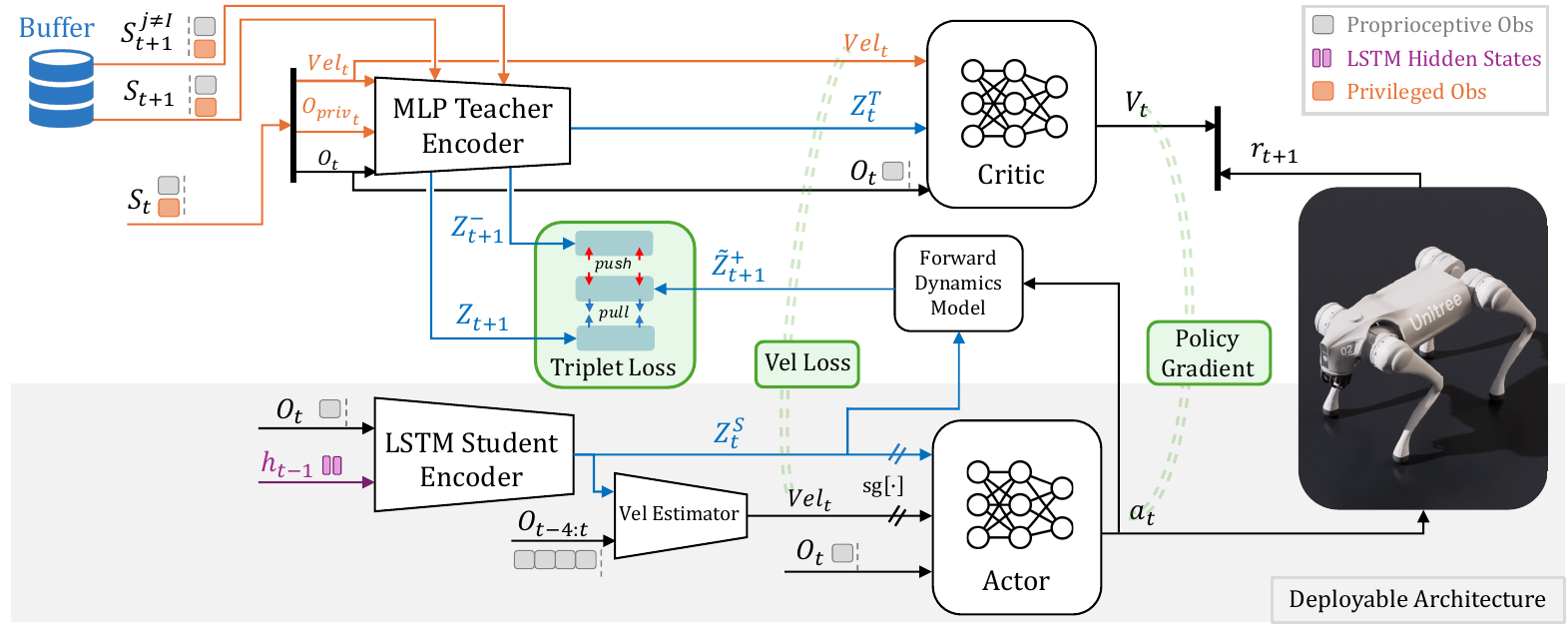} 
    \end{minipage}
  }
  \caption{The training framework includes a teacher encoder \(f_T\) that processes privileged states \(S\) to generate structured latent representations \(Z^T\). The student encoder \(f_S\) extracts proprioceptive features \(Z^S\) from observation \(O_{t}\) and hidden states \(h_{t-1}\). Our triplet loss pulls the student’s next-state prediction \( \tilde{Z}_{t+1}^{+}\) close to the teacher’s encoding \( Z_{t+1}\) and away from the teacher's encoding \( Z_{t+1}^-\) of other contexts sampled from the buffer. The policy gradient loss updates the actor and the critic, while the latter is also updated by the triplet loss. The velocity estimator's output is regressed with the ground truth velocity and is frozen after training to ensure future deployment adaptability}
  \label{fig:arch_full}
\end{figure*}

\subsection{Metric Learning}

Unlike reconstruction-based \gls{ssl}, \emph{metric learning} focuses on structuring the embedding space so that semantically similar samples lie close to each other while dissimilar samples are pushed apart. We discuss two major metric-learning approaches as follows:

\subsubsection{Prototypical Learning}

It treats each class as a “prototype” in the embedding space. During training, observations are projected onto a lower-dimensional embedding and then assigned to the nearest prototype. This encourages tighter clusters of semantically related embeddings and simplifies downstream decision processes by reducing redundancy.

Proto-RL \cite{yarats2021protorl} utilizes this approach to improve exploration efficiency, allowing \gls{rl} agents to quickly generalize over various states by referencing a set of learned prototypes. More recently, HIMLoco \cite{long2024him} implemented prototypical learning for quadruped locomotion, where historical and future observations are aligned to form meaningful prototypes using SwAV-style assignments \cite{caron2021swav} and Sinkhorn-Knopp optimization. While HiM incorporates privileged information within the critic, it does not fully integrate it into the representation learning, which, in our experiments, hindered the formation of a globally optimal embedding space.

\subsubsection{Contrastive Learning}

It is widely used in self-supervised contexts across domains (e.g., images, speech, and NLP) \cite{chopra2005contrastive}. It relies on the principle of comparing “positive” (similar) pairs against “negative” (dissimilar) pairs. In \gls{rl}, this can be achieved by treating consecutive states or augmented versions of the same observation as positives and unrelated states from a replay buffer as negatives. By maximizing similarity among positives and minimizing it among negatives, contrastive methods ensure that states with shared features remain close in the embedding space.

Popular implementations of contrastive learning include InfoNCE-based formulations \cite{oord2019contrastive} and SimCLR-style objectives \cite{chen2020simplecontrastive}. CURL \cite{srinivas2020curlcontrastive}, for example, employs the latter to achieve data-efficient \gls{rl} through image encoders. 

Another implementation is the triplet loss, which adopts a more explicit \emph{distance ordering} among three inputs: \emph{anchor} \(x^a\), \emph{positive} \(x^p\), and \emph{negative} \(x^n\). The objective mandates that the anchor be closer to the positive than the negative by at least a margin \(\alpha\). Formally:
\begin{equation}
    \label{eq:triplet_loss}
    \mathcal{L}_{\mathrm{triplet}} = \sum_{i=1}^{N} 
    \Big[
        \| f(x_i^a) - f(x_i^p) \|_2^2 
        - 
        \| f(x_i^a) - f(x_i^n) \|_2^2
        + 
        \alpha
    \Big]_+,
\end{equation}
where \([\cdot]_{+} = \max(0, \cdot)\) ensures the loss remains non-negative. 

SLR \cite{chen2024slr} applies triplet loss to quadruped control by encoding temporal dependencies, using the latent of the next observation as the anchor, the current history as the positive, and the randomly sampled latent as negative. This encourages smooth representation evolution but assumes that temporally adjacent states are inherently similar. HiMLoco \cite{long2024him} refines this by adopting a dual-encoder setup, where one encoder processes past observations and another encodes the next single observation, both to be pulled towards the nearest prototype to enhance temporal consistency.

\subsubsection{Challenges in Metric Learning}

A key challenge in contrastive learning is designing an efficient positive sampling strategy, as it directly impacts how well the latent space retains task-relevant information. In highly dynamic environments where stochastic transitions and unobservable external factors shape state evolution, assuming temporally adjacent states with the same modalities can introduce noisy positives, misaligned representations and degrading generalization.

Similarly, negative sampling plays a crucial role in enforcing representation separation. SLR \cite{chen2024slr} utilizes random negative sampling, where negatives are selected uniformly from the entire observation pool. This often yields uninformative or misleading negatives that provide insufficient semantic contrast for learning discriminative representations, failing to challenge the encoder to distinguish between meaningful variations.

\begin{figure*}[t]
  \centering
  \fboxsep=0pt 
  \fboxrule=0pt 
  \fbox{
    \begin{minipage}{0.9\textwidth} 
      \centering
      \includegraphics[width=\textwidth]{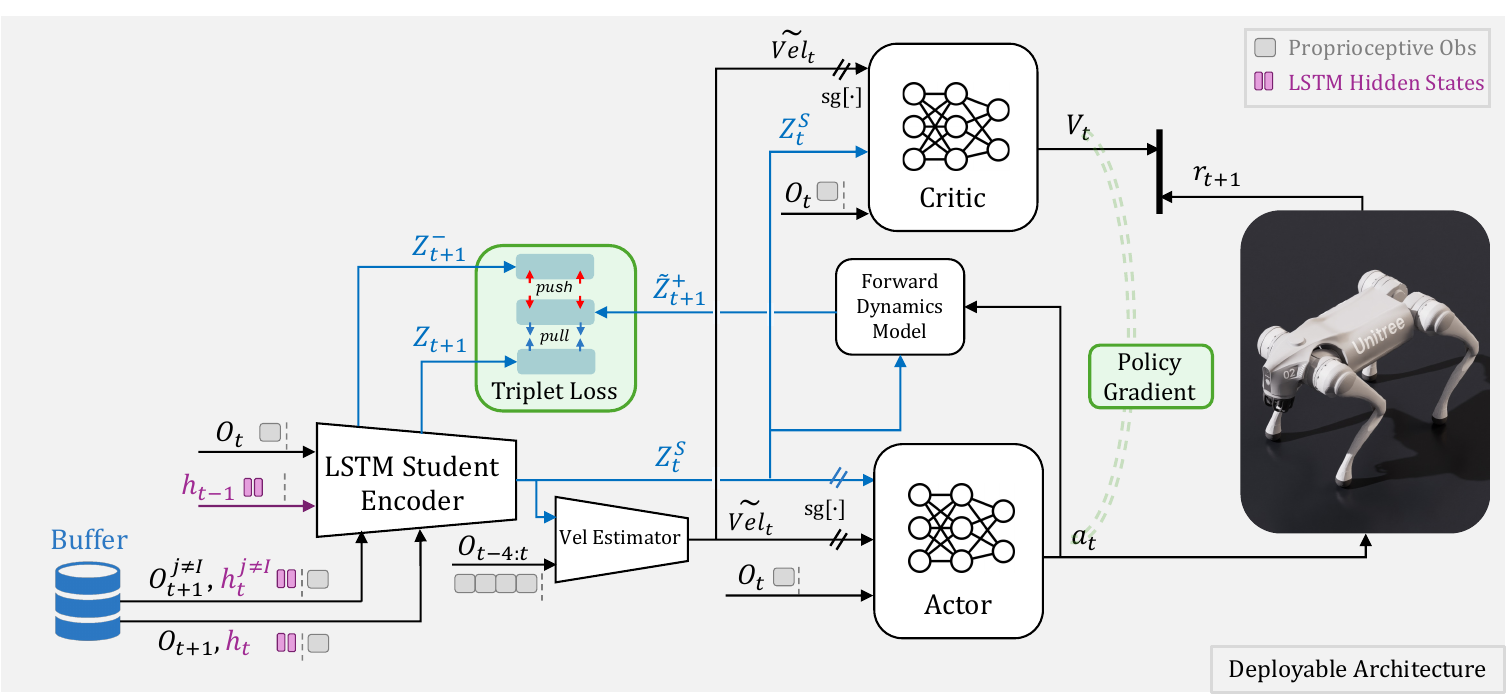} 
    \end{minipage}
  }
   \caption{During adaptation or privileged-free learning, the teacher encoder \( f_T \) is removed, and student encoder \( f_S \) constructs positive and negative sample pairs from the current agent’s proprioceptive observations \( O_{t+1} \) and those of another agent \( O_{t+1}^{j \neq i} \), along with their respective hidden states \( h_{t+1} \) and \( h_{t+1}^{j \neq i} \). This structured sampling enforces temporal consistency in the latent space, ensuring the student encoder learns meaningful representations without direct supervision. The absence of privileged teacher supervision makes the architecture inherently off-policy compatible and facilitates robust fine-tuning in dynamic and non-stationary environments.
   }
   \label{fig:arch_ft}
\end{figure*}

\subsection{Adaptive Continual Learning}

To address such limitations, we define the positive pairs from the teacher’s encoder utilizing privileged information rather than relying solely on temporal adjacency. Specifically, the teacher-informed latent of the next state serves as an anchor, while the actor’s predicted latent representation from proprioceptive history forms the positive sample as illustrated in Figure~\ref{fig:arch_full}. This structured sampling prevents representation collapse, allowing the student to infer task-relevant features without enforcing direct regression onto the teacher's latent.

For negative sampling, we constrain selections to states experienced by agents (trained in parallel under different environmental parameters) rather than drawing from a fully random pool. This approach enhances the semantic relevance of the learned representation by ensuring that the student’s prediction is pushed away from environmentally diverse negatives (e.g., with different friction, terrain, and payload) rather than arbitrary negatives. As a result, the encoder refines its feature space to implicitly encode unobservable but task-relevant properties, enabling the policy to distinguish meaningful proprioceptive patterns from underlying environmental dynamics that influence locomotion.

A key advantage of our design is its flexibility beyond privileged training. As illustrated in Figure~\ref{fig:arch_ft}, once deployed, the system could seamlessly transition to a proprioceptive-only regime by replacing the teacher's generated next-state latents (shown in Figure~\ref{fig:arch_full}) with self-predicted representations from the student encoder. This enables continual learning, enhancing sample efficiency and real-world adaptability in unstructured environments.
\section{Methodology}
\label{sec:methodology}

Inspired by self-distillation \cite{zhang2019byol} and contrastive learning \cite{chopra2005contrastive}, our method trains sample-efficient quadrupedal locomotion policies that generalize from simulation to reality through two phases: (1) simulation training with privileged information (Figure \ref{fig:arch_full}), and (2) real-world deployment with continual learning (Figure \ref{fig:arch_ft}).

\subsection{\acrlong{rl} Formulation}
The policy is trained using PPO \cite{schulman2017ppo}, which optimizes the policy \( \pi_{\theta} \) by maximizing the expected discounted return \(G_t = \mathbb{E} \Big[ \sum_{k=0}^{ K} \gamma^k r_k \Big]\), where \( \gamma \) is the discount factor, and \( r_t \) is the reward function. 

The state space \( s_t \) includes proprioceptive and privileged observations as follows: 
\begin{itemize}
    \item The policy network receives a proprioceptive-only input vector \(O_t\) with a recurrent hidden state \(h_{t-1}\), capturing historical proprioceptive states to enhance temporal dependencies. This includes base angular velocity \( \omega_t \in \mathbb{R}^{ 3} \), projected gravity \( g_t \in \mathbb{R}^{ 3} \), linear velocity commands \( v_t^{cmd} \in \mathbb{R}^{ 3} \), joint positions \( q_t \in \mathbb{R}^{ 12} \), joint velocities \( \dot{q}_t \in \mathbb{R}^{ 12} \), and previous actions \( a_{t-1} \in \mathbb{R}^{ 12} \).
    \item The critic network, utilizing privileged information \(O_{priv_t}\), receives an expanded input which includes all policy inputs at a single timestep \(t\), along with base linear velocity \( v_t \in \mathbb{R}^{1 \times 3} \), height scan \( h_t \in \mathbb{R}^{1 \times 187} \), base external force \( f_t^{ext} \in \mathbb{R}^{1 \times 3} \), foot contact states \( c_t \in \mathbb{R}^{1 \times 4} \), contact friction coefficient \( \mu_t \in \mathbb{R}^{1 \times 1} \), and payload mass \( m_t \in \mathbb{R}^{1 \times 1} \).
\end{itemize}

\begin{algorithm}[H]
\caption{PPO with \acrfull{ours} for Quadruped Locomotion}
\label{alg:ppo_ours}
\begin{algorithmic}
\REQUIRE Randomly initialize policy \( \pi_{\theta} \), value function \( V_{\phi} \), teacher, student encoders \( f_T \) and \( f_S \), forward dynamics model \( f_D \), velocity estimator \( f_V \), and replay buffer \( \mathcal{B} \)

\FOR{\( 0 \leq \text{iter} \leq N_{iter}^{total} \)}
    \FOR{\( 0 \leq t \leq T \)}
        \STATE \( o_t, s_t \leftarrow \text{Observe} \)
        \STATE \( Z_t^a = f_S(o_{t}, h_{t-1}) \)
        \STATE \( \hat{v} = f_V(Z_t^a, o_{t-3:t}) \)
        \STATE \( a_t \sim \pi_{\theta}(a_t | o_t, Z_t^a, \hat{v}) \)
        \STATE \( o_{t+1}, s_{t+1}, r_t \leftarrow \text{env.step}(a_t) \)
        \STATE Store \( (o_t, h_{t-1}, s_t, a_t, r_t, o_{t+1}, h_{t}, s_{t+1}) \) in \( \mathcal{B} \)
    \ENDFOR

    \FOR{\( k \leq N_{\text{updates}} \)}
        \STATE Sample random mini-batch from \( \mathcal{B} \)
        \STATE \( Z_{t+1} = f_T \left( S_{t+1} \right) \) 
        \STATE \( \tilde{Z}_{t+1}^{+} = f_D \left( f_S(O_{t}, h_{t-1}), a_t \right) \) 
        \STATE \( Z_{t+1}^{-} = f_S \Big( O_{t-h+1:t}^{j} \Big), \quad j \sim \text{Uniform}(\mathcal{B} \setminus \mathcal{T}_i) \) 
        \vspace{0.5em}
        \STATE Compute \( \mathcal{L}_{\text{PPO}} \), \( \mathcal{L}_{\text{triplet}}\) and \( \mathcal{L}_{\text{vel}} \) (Eq. 2-4 and  \cite{schulman2017ppo}).
        \vspace{-0.5em}
        \STATE Update:
        \vspace{-1.9em}
        \[
        \theta_T \leftarrow \theta_T - \lambda \nabla_{\theta_T} \left( \mathcal{L}_{\text{value}} + \mathcal{L}_{\text{triplet}} \right)
        \]
        \vspace{-1.25em}
        \[
        \theta_S \leftarrow \theta_S - \lambda \nabla_{\theta_S} \left( \mathcal{L}_{\text{vel}} + \mathcal{L}_{\text{triplet}} \right)
        \]
        \vspace{-1.25em}
        \[
        \theta_D \leftarrow \theta_D - \lambda \nabla_{\theta_D} \mathcal{L}_{\text{triplet}}
        \]
        \vspace{-1.25em}
        \[
        \theta, \phi \leftarrow \theta, \phi - \lambda \nabla_{\theta, \phi} \mathcal{L}_{\text{PPO}}
        \]
        \vspace{-1.75em}
    \ENDFOR
    \STATE Empty \( \mathcal{B} \)
\ENDFOR
\end{algorithmic}
\end{algorithm}

The action space is defined as \( a_t \in \mathbb{R}^{12} \), representing the target joint torques applied to the actuators. The training hyperparameters, reward function, and domain randomization ranges are listed in Tables~\ref{tab:hyperparams}, \ref{tab:rewards}, and \ref{tab:domain_randomization} in Appendix \ref{appendix}.


\begin{figure*}[t!]
     \centering

    \includegraphics[width=0.8\textwidth]{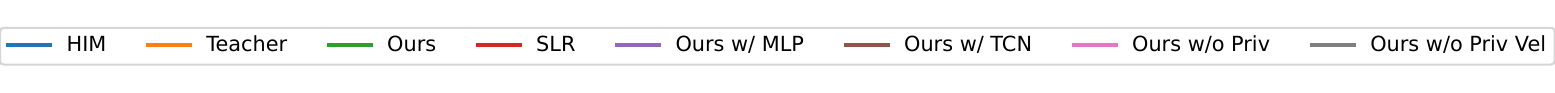}
    \vspace{-0.2em}

    \begin{minipage}[t]{0.33\textwidth}
        \includegraphics[width=\linewidth]{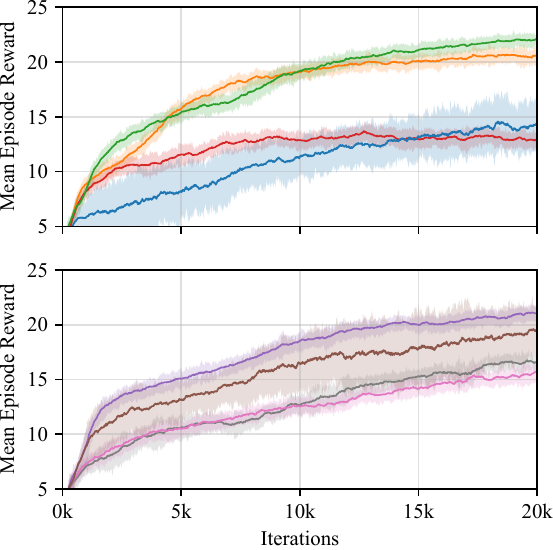}
    \end{minipage}%
    \begin{minipage}[t]{0.33\textwidth}
        \includegraphics[width=\linewidth]{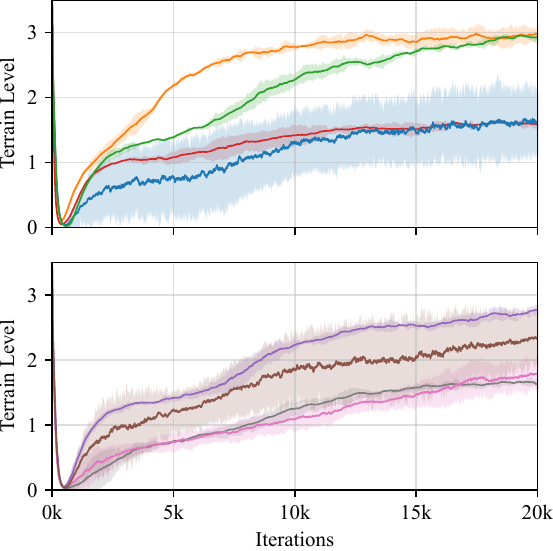}
    \end{minipage}%
    \begin{minipage}[t]{0.33\textwidth}
        \includegraphics[width=\linewidth]{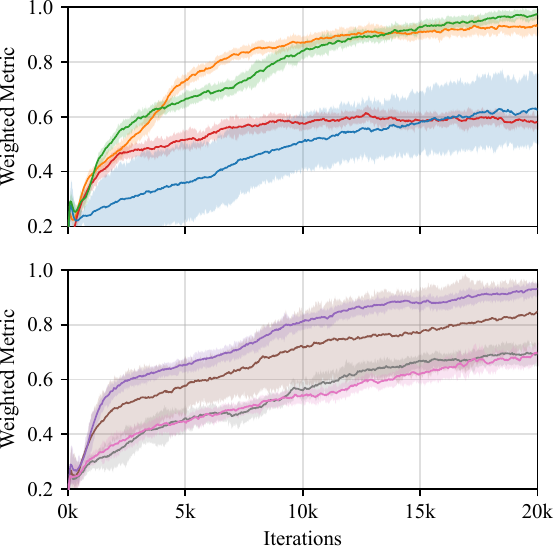}
    \end{minipage}
    \caption{Training results of baseline algorithms and our model variants across three seeds. [Left]: Training reward, [Middle]: Terrain level, and [Right]: Our weighted performance metric, computed as:  
    $M_{\text{train}} = 0.25 \times \text{Normalized Terrain Level} + 0.6 \times \text{Normalized Mean Reward} + 0.15 \times \text{Normalized Episode Length}$.}
    \label{fig:train_plots}
\end{figure*}


\subsection{Training Framework with Privileged Information}
Our \gls{ours} approach leverages privileged information from the \emph{teacher encoder}, while the \emph{student encoder} and \textit{actor network} operate exclusively on proprioceptive observations. Figure~\ref{fig:arch_full} illustrates the architecture as follows::

\subsubsection{PPO-Based Policy Optimization}
The actor and critic networks are trained using the PPO objective:
\begin{equation}
    \mathcal{L}_{\text{PPO}} = \mathbb{E} \Big[ \min( r_t(\theta) A_t, \text{clip}(r_t(\theta), 1 - \epsilon, 1 + \epsilon) A_t ) \Big],
\end{equation}

where \( r_t(\theta) \) is the probability ratio between the new and old policies, and \( A_t \) is the advantage estimate computed via generalized advantage estimation (GAE) \cite{schulman2017ppo}.

The actor policy is conditioned on student latent and velocity estimates \(a_t \sim \pi(a_t | O_t, Z_t^S, \hat{vel}_t)\), where \( \hat{vel}_t \) is the estimated base velocity obtained from a learned model \(\hat{vel}_t = f_V(Z_t^S, O_{t-4:t})\). The velocity estimator utilizes a 4-step history to capture temporal dependencies inspired by \cite{li2024bipedalversatile}.
During training, the velocity estimator is optimized using a mean squared error (MSE) loss:
\begin{equation}
    \mathcal{L}_{\text{vel}} = \mathbb{E} \Big[ \| \hat{v}_t - v_t^{\text{true}} \|^2 \Big].
\end{equation}
This loss minimizes the prediction error between the estimated velocity \( \hat{v}_t \) and the ground-truth velocity \( v_t^{\text{true}} \). To prevent interference with policy learning, we apply \(sg[\cdot]\) operator to prevent the estimator’s gradients from propagating into the policy, enabling modular and stable learning.

\subsubsection{Contrastive Representation Learning with Triplet Loss}
Rather than directly regressing the student latent \( Z_t^S \) onto the teacher latent \( Z_t^T \), we employ a contrastive triplet loss formulation to enforce structured representation learning:
\vspace{-1em}

\begin{equation}
    \mathcal{L}_{\text{triplet}} = \sum_{i=1}^{N} \Big[
    \| f(Z_i^T) - f(Z_i^+) \|_2^2
    - 
    \| f(Z_i^T) - f(Z_i^-) \|_2^2
    + 
    \alpha
    \Big]_+,
\end{equation}

where the \textbf{anchor} \( Z_{t+1}^T \) represents the teacher’s next latent, the \textbf{positive} sample \( \tilde{Z}_{t+1}^+ \) corresponds to the student’s predicted next latent from the dynamics model, and the \textbf{negative} sample \( Z_{t+1}^- \) is a latent drawn from rollouts under different environmental conditions.

This formulation ensures that the student’s next-state prediction is pulled towards the teacher’s rich, informative latent representation while being pushed away from other sampled trajectories. As these trajectories were subjected to domain-randomized environmental variations (e.g., altered friction, mass, and perturbations), the triplet loss encourages the student’s state encoding to be both task-relevant and robust.

\subsubsection{Student and Critic Encoder Optimization}
The critic encoder is optimized jointly with the PPO value loss and triplet loss \(
    \mathcal{L}_{\text{critic}} = \mathcal{L}_{\text{value}} + \lambda_{\text{triplet}} \mathcal{L}_{\text{triplet}}\).
The student encoder is updated using only gradients from the forward dynamics model \(\theta_S \leftarrow \theta_S - \lambda \nabla_{\theta_S} \mathcal{L}_{\text{triplet}}\).
To prevent representation collapse, gradient flow from the actor policy is blocked from propagating into both the student encoder and the velocity estimator. The full training pipeline is summarized in Algorithm \ref{alg:ppo_ours}.

\subsection{Deployable Architecture for Real-World Fine-Tuning}
To facilitate continual learning in the real world without privileged observations, we remove the teacher encoder and generate positive and negative samples exclusively from the student encoder (Figure~\ref{fig:arch_ft}).
The positive pair is now defined as \(Z_{t+1}^+ = f_S(O_{t-h+1:t+1})\), whereas the negative sample is drawn from past rollouts \(Z_{t+1}^- = f_S(O_{t-h+1:t+1}^{j}), \quad j \neq I\).

This enables real-world fine-tuning without reliance on privileged exteroceptive information, making it compatible with off-policy RL algorithms such as Soft Actor Critic (SAC) \cite{haarnoja2018sac}.

\section{Results and Discussion}
\label{sec:results}

We trained our policy using NVIDIA Isaac Sim (Figure~\hyperref[fig:terrain_in_sim]{2}) and deployed it on Unitree Go2 (Figure~\ref{fig:deployment}) to evaluate performance across simulation and real-world environments.

\subsection{Training Performance Analysis}

To evaluate the effectiveness of our proposed method, we trained our algorithm alongside four of its variants and three state-of-the-art baselines, namely \emph{\textbf{HIM}} \cite{long2024him}, \emph{\textbf{SLR}} \cite{chen2024slr}, and a \emph{\textbf{privileged Teacher}}. The \emph{Teacher} serves as an expert model with full access to privileged information in both the actor and critic networks. Its architecture was selected from several candidates—including MLP and RNN encoders—to maximize performance with privileged inputs. The variants of our model include:
\begin{itemize}
    \item \textbf{\emph{Ours w/ MLP}}: A 10-step MLP student encoder.
    \item \textbf{\emph{Ours w/ TCN}}: A Temporal Convolutional Network (TCN) with hidden channels [32, 32, 32], kernel sizes [8, 5, 5], and strides [4, 1, 1] \cite{peng2024bipedawalk}.
    \item \textbf{\emph{Ours w/o Priv}}: The same architecture but trained without privileged information, as in Figure \ref{fig:arch_ft}.
    \item \textbf{\emph{Ours w/o Priv Vel}}: Similar to the previous variant but without access to velocity information in both actor and critic, making it comparable to \emph{SLR} while incorporating our proposed negative sampling method.
\end{itemize}

All models were trained for 20,000 iterations, with evaluations conducted every 2,500 iterations to monitor the trade-off between underfitting and overfitting. To provide a unified training performance metric, we employ a weighted combination of normalized terrain levels, mean reward, and episode length, with respective weights of 0.25, 0.6, and 0.15.

Figure~\ref{fig:train_plots} illustrates the training progression based on this metric, averaged over three random seeds. Our method consistently achieved the highest performance, surpassing the \emph{Teacher} after 12k iterations while achieving 4.1\% higher return. The \emph{MLP} variant achieved comparable results, reaching 95.5\% of our final performance. While \emph{HIM} showed less robustness, it improved steadily and surpassed \emph{SLR} after 15k iterations, ultimately reaching 63.9\% of our performance. \emph{SLR} was the most stable but achieved the lowest return at 58.7\%. Full results and training curves are available \href{https://amrmousa.com/TARLoco/}{online}.

\subsection{Generalization and Robustness Evaluation}

\begin{figure}[t]
    \centering
    \includegraphics[width=\linewidth]{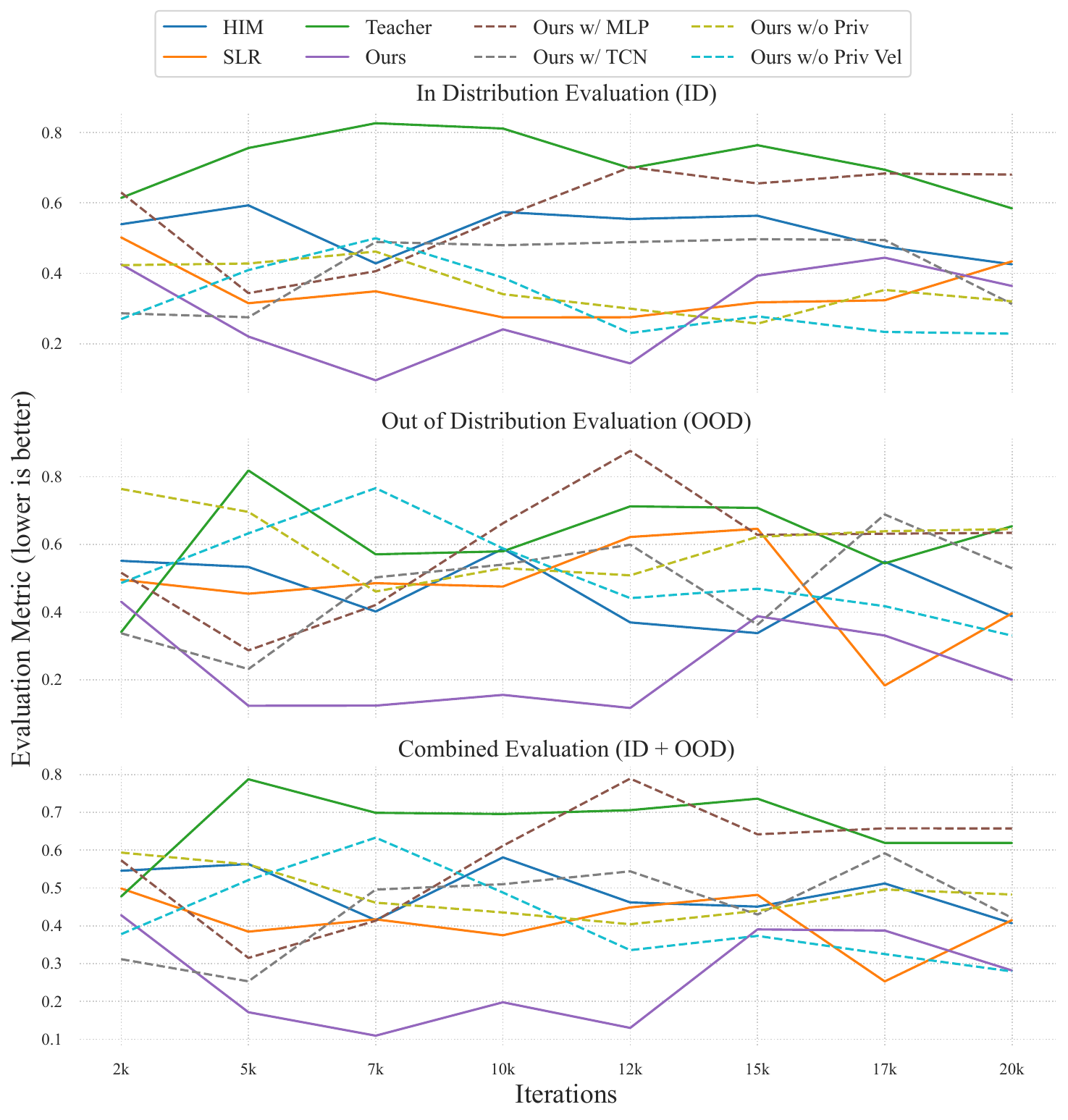} 
    \caption{Evaluation results of all models across \gls{id} and \gls{ood} settings.}
    \label{fig:eval_plots}
\end{figure}

We evaluated generalization using a composite metric across both \gls{id} and \gls{ood} test cases. The metric combines normalized mean linear and angular velocity errors, along with the failure counts (falls), weighted at 0.3, 0.15, and 0.55, respectively. Each component is normalized over the observed range across all methods to ensure fair comparison.

\textbf{\gls{id} test cases} include variations within the training domain: friction levels \([0.1, 1.0]\), payload masses \([0, 7.5]~\text{kg}\), and a maximum linear velocity of \( 1.0~\text{m/s} \).  \textbf{\gls{ood} test cases} introduce conditions outside the training range, such as a payload mass of \( 15~\text{kg} \) and an increased maximum linear velocity of \( 2.0~\text{m/s} \).  

Figure~\ref{fig:eval_plots} presents the evaluation results. Our algorithm consistently outperformed all other variants, particularly in \gls{ood} scenarios, achieving peak performance at 7,500 iterations with a 74\% lower combined error than the expert \emph{Teacher}, 39.8\% and 42.2\% lower than \emph{SLR} and \emph{HIM}, respectively. While models utilizing MLP and TCN architectures peaked at 5,000 iterations, our method continued improving until 7,500, suggesting a possible earlier termination criterion for efficient training.

Other baselines, including \emph{HIM}, peaked at 12,500 iterations, while \emph{SLR} required up to 17,500 iterations to reach its best performance—still 22.1\% less in performance than our model at just 7,500 iterations. Our approach consistently outperformed all models across different training durations, maintaining high performance from 5,000 to 12,500 iterations without degradation.

We quantified the contribution of each component in our algorithm using the combined evaluation across both \gls{id} and \gls{ood} settings. Leveraging privileged information led to the largest performance gain of 28.2\%, followed by 8.0\% from improved negative sampling, 3.8\% from incorporating a recurrent policy, and 0.82\% from velocity estimation.

\begin{figure*}[t]
    \centering
    \includegraphics[width=0.34\linewidth]{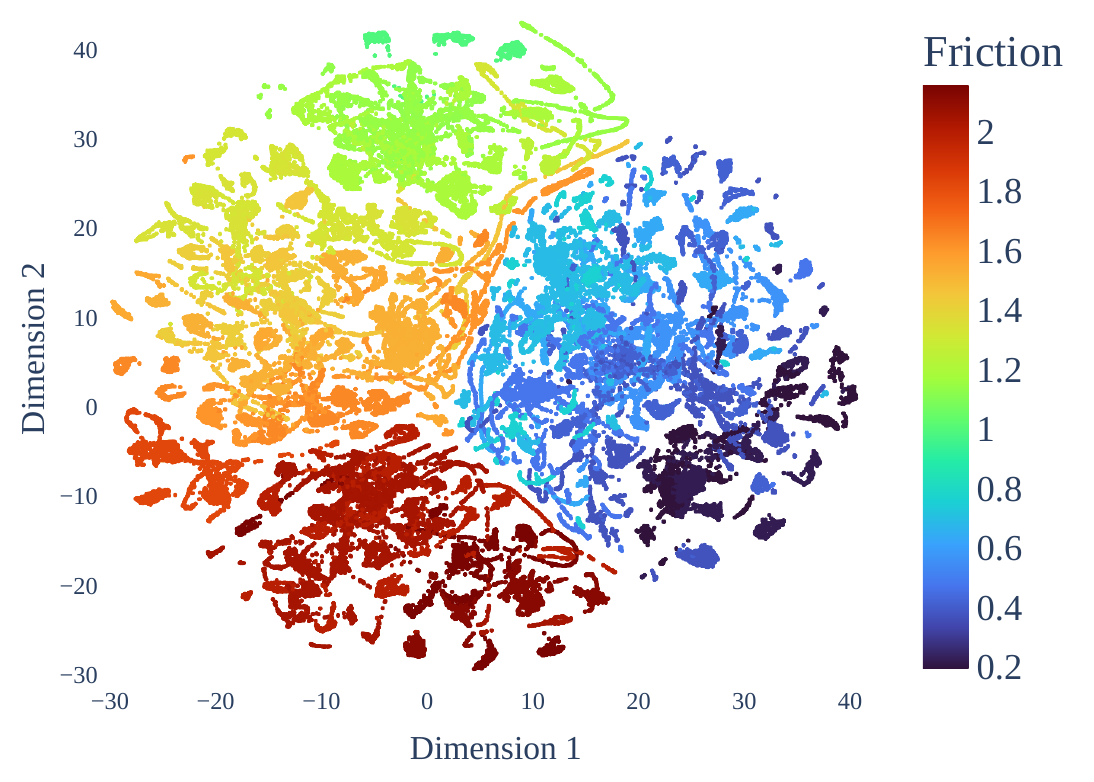}\hspace{-7.5pt}%
    \includegraphics[width=0.34\linewidth]{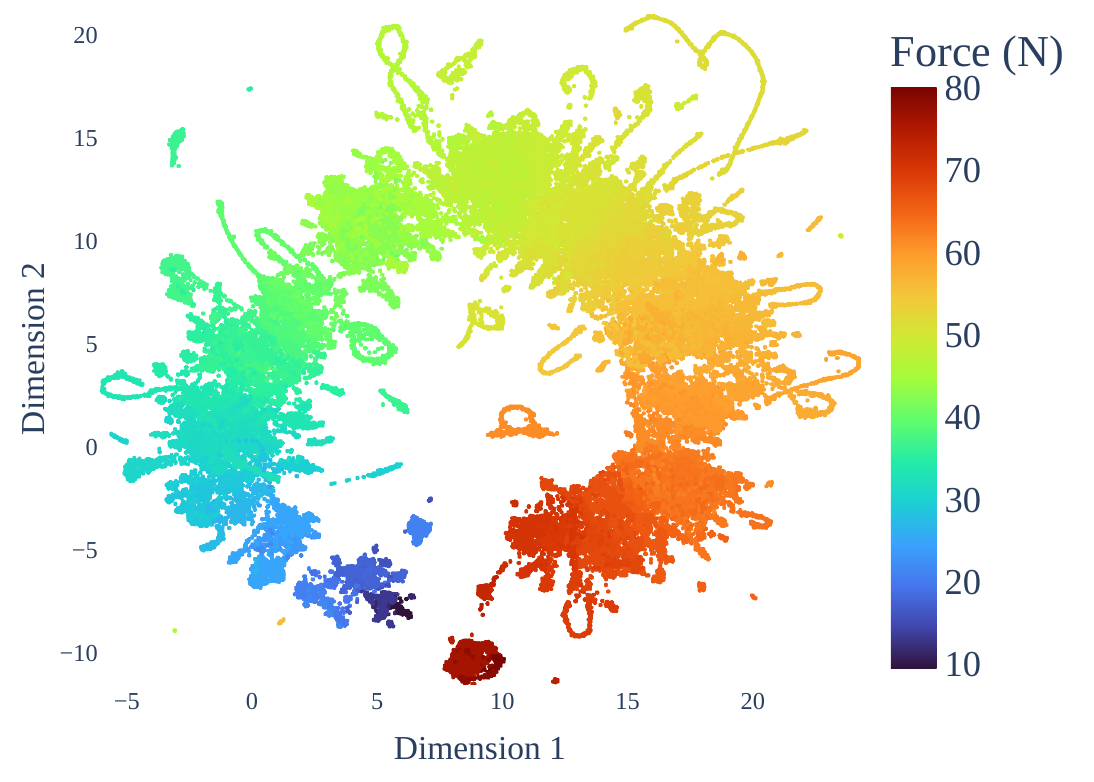}\hspace{-7.5pt}%
    \includegraphics[width=0.34\linewidth]{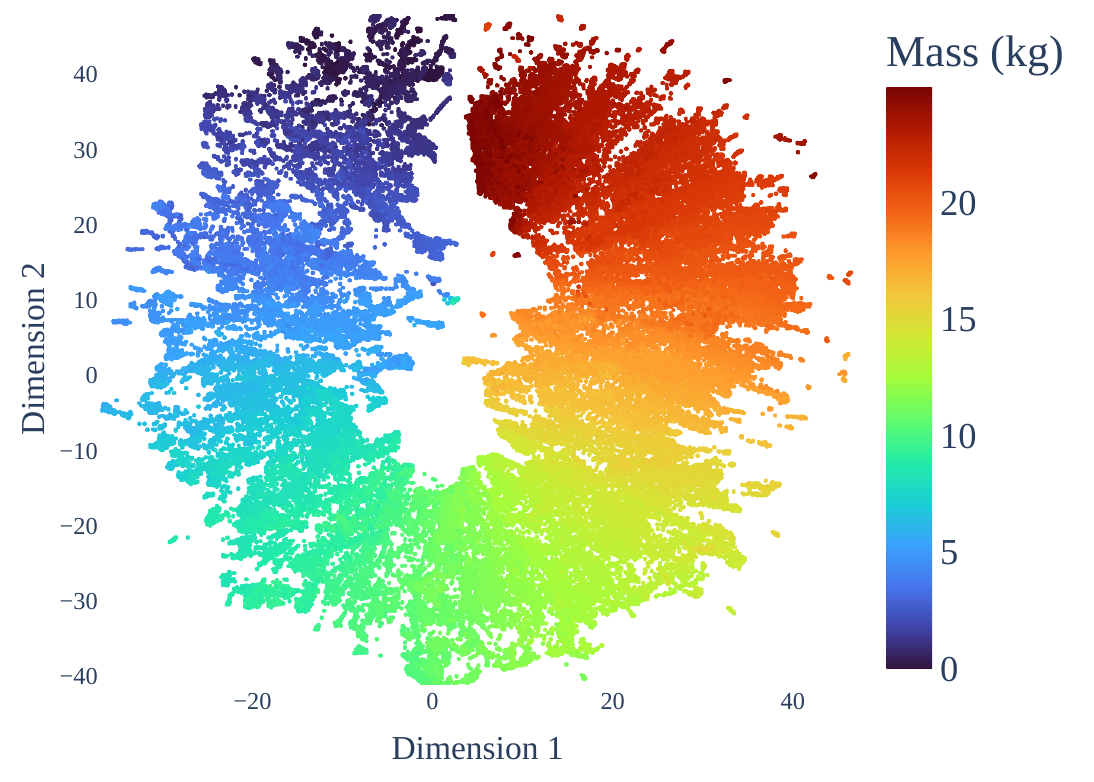}
    \caption{T-SNE projections of the student encoder’s latent representations under isolated variation of friction (left), external force (center), and mass (right). Each factor was varied independently beyond the training distribution to evaluate the encoder’s inference capability.}
    \label{fig:T-SNE_all}
\end{figure*}

\subsection{Impact of Privileged Information and Representation Learning}

A key observation from our experiments is the significant performance gap in \gls{ood} settings between methods that leverage privileged information during training and those that do not. The largest performance degradation was observed in \emph{SLR}, \emph{Ours w/o Priv}, and \emph{Ours w/o Priv Vel}, indicating that privileged information is crucial for training robust policies that generalize to unseen scenarios. However, models trained exclusively with privileged information, such as the \emph{1-step MLP Teacher}, failed to outperform our approach, suggesting that representation learning is critical in encoding environment dynamics and temporal dependencies.

Our method balances the strengths of privileged information and self-supervised learning by leveraging contrastive alignment from a privileged teacher encoder to extract meaningful representations. This approach improves training efficiency and generalization while preserving adaptability for real-world deployment.

\subsection{Encoder t-SNE Visualization}

Figure~\ref{fig:T-SNE_all} shows t-SNE visualizations of the student encoder’s latent space under isolated variation of friction, external force, and payload—each tested beyond training randomization ranges. 

In each case, the latent representations form a ring-like manifold encoding both locomotor phase and gradual variation in the extrinsic parameters. 
This structured geometry demonstrates that the encoder preserves the underlying periodicity of the gait while remaining sensitive to meaningful physical variations by learning physics-grounded embeddings rather than merely memorizing training distributions.

\subsection{Hardware Validation and Robustness Evaluation}

Our policy $\pi_{\theta}$ was deployed on a Unitree Go2, as shown in Figure~\ref{fig:deployment}, and evaluated through robustness tests including:
(i) a diverse set of terrains such as slippery flooring, high-friction rubber mats, bouncy deformable foam, coarse asphalt, and dense vegetation;
(ii) payload mass was varied in the range 0–12 kg, covering multiple configurations up to the robot’s mechanical limit;
(iv) an \gls{ood} actuator degradation test, reducing the commanded torque of one joint to 10\% of the nominal value to emulate hardware failure;
(iii) while trotting at 1.0 m/s, the robot absorbed lateral pushes up to 150 N and navigated vertical steps from +30 cm to -60 cm.

The policy consistently completed all trials with stable performance, avoiding joint torque saturation and emergency shutdowns. In contrast, the vendor-provided baseline controller showed degraded performance and was unable to handle several of these scenarios. Our robot also demonstrated agile bidirectional stepping and maintained stability with partial leg obstruction or actuator degradation. These results highlight strong sim-to-real generalization in a zero-shot setting. Full experimental details and supporting videos are available on our \href{https://amrmousa.com/TARLoco/}{project website}.
\section{Conclusion}
\label{sec:conclusion}
We presented a \acrfull{ours} approach for quadrupedal locomotion, addressing the challenges of privileged representation misalignment and real-world generalization. By leveraging privileged information in simulation to construct structured latent spaces while employing contrastive objectives for robust representation learning, our method achieves superior performance in both training and generalization.

Experimental results demonstrate that our method surpasses state-of-the-art baselines in both \gls{id} and \gls{ood} evaluation settings. It achieves optimal performance significantly earlier in training while maintaining strong generalization. Our model exhibits greater sample efficiency, requiring only 7,500 iterations to outperform models trained for 20,000 iterations.

Furthermore, our ablation studies highlight the necessity of privileged information for robust policy learning and confirm that self-supervised representation learning improves adaptability, training stability, and long-term generalization. Unlike purely privileged-based approaches, our method retains the ability to fine-tune or even train from scratch in real-world deployments, making it well-suited for long-term autonomous adaptation.

Future work will explore extending our framework to off-policy \gls{rl} paradigms, integrating fine-tuning strategies for real-world continual learning, and evaluating performance across diverse robotic morphologies.

\bibliographystyle{IEEEtran}
\bibliography{bibliography} 
\section{APPENDIX}
\label{appendix}

\subsection{Hyperparameters}
\label{app:sec:hyperparams}
The hyperparameters used during training are shown in Table~\ref{tab:hyperparams}.

\small 

\begin{table}[h]
\centering
\caption{Hyperparameters}
\begin{adjustbox}{max width=\linewidth}
\begin{tabular}{p{2.4cm} p{1.7cm} p{2.0cm} p{1.8cm}}
\toprule
\textbf{Param} & \textbf{Value} & \textbf{Param} & \textbf{Value} \\
\midrule
Optimizer & Adam & Adaptive lr & 5e-5 - 1e-3 \\
Gamma $\gamma$ & 0.99 & Lambda $\lambda$ & 0.95 \\
Triplet Loss Coef. & 1.0 & KL Coef. & 0.01 \\
Mini-batches & 4 & Num of Epochs & 5 \\
Activation Func. & ELU & Latent Dim. & 45 \\
LSTM Enc. & [256] & Actor/Critic & [512, 256, 128] \\
Dynamics Model & [64] \\
\bottomrule
\end{tabular}
\end{adjustbox}
\label{tab:hyperparams}
\end{table}

\subsection{Reward Functions}
\label{app:sec:rewards}

The reward terms used during training are presented in Table~\ref{tab:rewards}, following the definition of \cite{kumar2021rma, margolis2023walk}.

\begin{table}[h]
\centering
\caption{Reward Terms}
\begin{adjustbox}{max width=\linewidth}
\begin{tabular}{p{2.5cm} p{1.5cm} p{2.0cm} p{1.8cm}}
\toprule
\textbf{Term} & \textbf{Value} & \textbf{Term} & \textbf{Value} \\
\midrule
Lin Vel (xy) Exp. & 1.5 & Ang Vel (z) Exp. & 0.75 \\
Lin Vel (z) & -2.0 & Ang Vel (xy) & -0.05 \\
Joint Torque & -0.0002 & Joint Accel. & -2.5e-7 \\
Action Rate & -0.01 & Feet Air Time & 0.01 \\
Undesired Contacts & -1.0 & & \\
\bottomrule
\end{tabular}
\end{adjustbox}
\label{tab:rewards}
\end{table}

\subsection{Domain Randomization}
\label{app:sec:domainrand}
The domain randomization parameters used during training are listed in Table~\ref{tab:domain_randomization}, following the methodology from \cite{wu2023amp, nahrendra2023dreamwaq}.

\begin{table}[h]
\centering
\caption{Domain Randomization}
\begin{adjustbox}{max width=\linewidth}
\begin{tabular}{p{2.1cm} p{1.8cm} p{2.1cm} p{1.8cm}}
\toprule
\textbf{Param} & \textbf{Range} & \textbf{Param} & \textbf{Range} \\
\midrule
Friction & [0.1, 3.0] & Restitution & [0.0, 1.0] \\
Payload Range & [-2, 10] (kg) & Ext. Force & $\pm20$ (N)\\
Ext. Torque & $\pm5$ (N.m) & Joint Init. Pos. & [0.5, 1.5]\textsuperscript{*} \\
\bottomrule
\end{tabular}
\end{adjustbox}
\label{tab:domain_randomization}
\end{table}

\vspace{-4mm}
\noindent{\scriptsize\textsuperscript{*} Scaling values (multipliers of nominal values).}

\end{document}